\documentclass[letterpaper]{article} % DO NOT CHANGE THIS
\usepackage{aaai25}  % DO NOT CHANGE THIS
\usepackage{times}  % DO NOT CHANGE THIS
\usepackage{helvet}  % DO NOT CHANGE THIS
\usepackage{courier}  % DO NOT CHANGE THIS
\usepackage[hyphens]{url}  % DO NOT CHANGE THIS
\usepackage{graphicx} % DO NOT CHANGE THIS
\usepackage{natbib}  % DO NOT CHANGE THIS AND DO NOT ADD ANY OPTIONS TO IT
\usepackage{caption} % DO NOT CHANGE THIS AND DO NOT ADD ANY OPTIONS TO IT
\usepackage{algorithm}
\usepackage{algorithmic}

\usepackage{newfloat}
\usepackage{listings}
\DeclareCaptionStyle{ruled}{labelfont=normalfont,labelsep=colon,strut=off} % DO NOT CHANGE THIS
\floatstyle{ruled}
\newfloat{listing}{tb}{lst}{}
\floatname{listing}{Listing}
\usepackage{amsmath}
\usepackage{amssymb}
\usepackage[table,xcdraw]{xcolor}
\usepackage{multirow}
\usepackage{makecell}
\usepackage{pifont}
\usepackage{arydshln}
\newcommand*{\ourmethod}{\textit{TeamLoRA}}
\usepackage{todonotes}
\usepackage{fontawesome}
\usepackage{bm}
\title{\includegraphics[height=0.3in, width=0.3in]{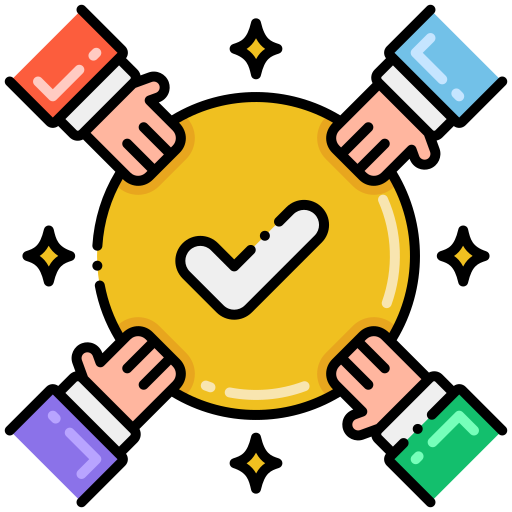} TeamLoRA: Boosting Low-Rank Adaptation with Expert Collaboration and Competition}
\author {
    \small
    Tianwei Lin\textsuperscript{\rm 1},
    Jiang Liu\textsuperscript{\rm 1},
    Wenqiao Zhang\textsuperscript{\rm 1},
    Zhaocheng Li\textsuperscript{\rm 1},
    Yang Dai\textsuperscript{\rm 1},
    Haoyuan Li\textsuperscript{\rm 2},
    Zhelun Yu\textsuperscript{\rm 2},
    Wanggui He\textsuperscript{\rm 2},
    Juncheng Li\textsuperscript{\rm 1},
    Hao Jiang\textsuperscript{\rm 2},
    Siliang Tang\textsuperscript{\rm 1},
    Yueting Zhuang\textsuperscript{\rm 1},
}
\affiliations {
\small
    \textsuperscript{\rm 1}Zhejiang University\\
    \textsuperscript{\rm 2}Alibaba Group\\
    \{tianweilin, jiangliu, wenqiaozhang, zhaochengli, yangdai, junchengli, siliang, yzhuang\}@zju.edu.cn,\\
    \{lihaoyuan.lhy, yuzhelun.yzl, wanggui.hwg, aoshu.jh\}@alibaba-inc.com
}
\begin{document}

\maketitle

\begin{abstract}
While Parameter-Efficient Fine-Tuning (PEFT) methods like LoRA have effectively addressed GPU memory constraints during fine-tuning, their performance often falls short, especially in multidimensional task scenarios. To address this issue, one straightforward solution is to introduce task-specific LoRA modules as domain experts, leveraging the modeling of multiple experts' capabilities and thus enhancing the general capability of multi-task learning.
Despite promising, these additional components often add complexity to the training and inference process, contravening the efficient characterization of PEFT designed for. Considering this, we introduce an innovative PEFT method, \includegraphics[height=0.12in, width=0.12in]{fig/cooperation.png} \textbf{\ourmethod{}}, consisting of a collaboration and competition module for experts, and thus achieving the right balance of effectiveness and efficiency:
(i) For \emph{collaboration}, a novel knowledge-sharing and -organizing mechanism is devised to 
appropriately reduce the scale of matrix operations, thereby boosting the training and inference speed.
% appropriately reduce the parameter scale, thereby boosting the training and inference speed.
(ii) For \emph{competition}, we propose leveraging a game-theoretic interaction mechanism for experts, encouraging experts to transfer their domain-specific knowledge while facing diverse downstream tasks, and thus enhancing the performance.
By doing so, \ourmethod{} elegantly connects the experts as a ``\textit{Team}'' with internal collaboration and competition, enabling a faster and more accurate PEFT paradigm for multi-task learning. 
To validate the superiority of \ourmethod{}, we curate a \texttt{comprehensive multi-task evaluation} (CME) benchmark to thoroughly assess the capability of multi-task learning. Experiments conducted on our CME and other benchmarks indicate the effectiveness and efficiency of \ourmethod{}. Our project is available at https://github.com/Lin-Tianwei/TeamLoRA.
\end{abstract}

\section{Introduction}
% In the last few years, Large Language Models (LLMs) pre-trained on ultra large-scale
% data have seen a great surge of interest with promising performance
% Large Language Models (LLMs)~\cite{touvron2023llama,anil2023palm,achiam2023gpt,young2024yi,reid2024gemini,cai2024internlm2,yang2024qwen2} 
% offers a versatile and task-agnostic foundation that underpins an extensive array of applications, derived from their knowledge storage and exceptional instruction-following capabilities. 
% have exhibited remarkable potential across various natural language application scenarios, renowned for their robust knowledge storage and exceptional instruction-following capabilities. 

\begin{figure}
    \centering
    \includegraphics[width=0.95\linewidth]{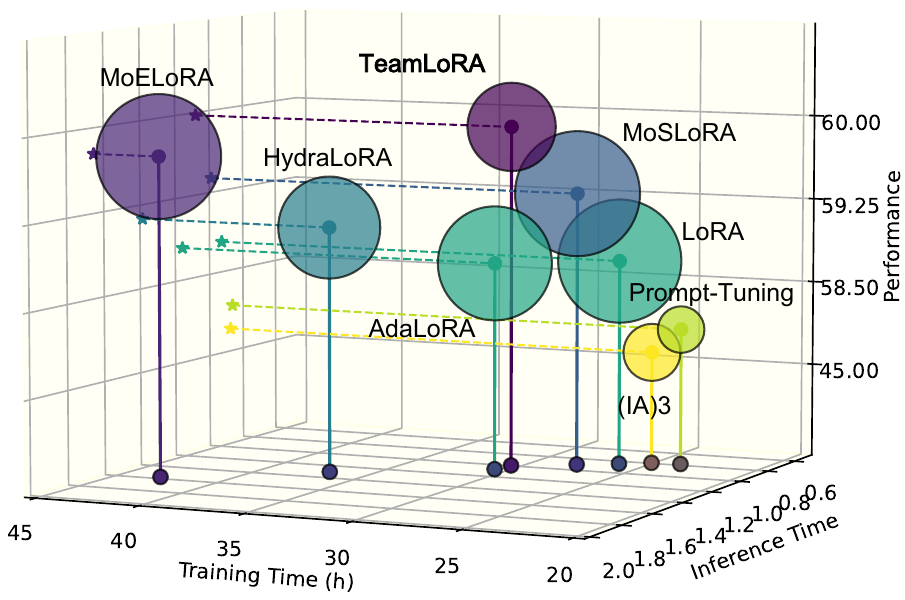}
    \caption{{Visualization of training time, inference time and performance for various PEFT methods on the CME benchmark.} The radius of the sphere illustrates the relative parameter scale added by different methods.}
    \label{fig:Overall of PEFT Methods Comparison}
\end{figure}

Instruction fine-tuning of Large Language Models 
(LLMs)~\cite{achiam2023gpt,reid2024gemini,cai2024internlm2,yang2024qwen2} and Multimodal Large Language Models (MLLMs)~\cite{ref:CLIP, li2022blip, huang2023language, ref:GPT4, zhang2024hyperllava} has achieved impressive proficiency in Natural Language Processing (NLP) and multi-modal understanding by effectively adapting \emph{task-agnostic} foundations to \emph{task-specific} domains. However, this approach requires substantial memory and computational resources for full fine-tuning (FFT), \emph{i.e.}, fine-tuning models with more than one billion parameters, which hinders its applicability. Therefore, Parameter-Efficient Fine-Tuning (PEFT)  techniques have emerged with the aim of reducing the cost by fine-tuning a small subset of parameters,  offering a streamlined approach for domain adaptation. 
% of fine-tuning by freezing pre-trained weights or updating smaller auxiliary modules in resource-constrained environments.
Among these methods, Low-Rank Adaptation (LoRA)~\cite{hu2022lora}, a popular PEFT approach, fine-tunes models by adapting lightweight auxiliary modules \( \Delta W = AB \) on top of pre-trained weights \( W_0 \), where $A$ and $B$ are low-rank matrices. LoRA offers performance comparable to full fine-tuning when focusing on the \textit{one-dimensional} domain or task with less computational effort. Nonetheless, qualitative research highlights LoRA's limitations in handling \textit{multidimensional} task scenarios, mainly due to the catastrophic forgetting and interference~\cite{kalajdzievski2024scalinglawsforgettingfinetuning} between tasks in the training stage.

One straightforward solution is to adaptively integrate the knowledge diversity of multiple LoRA experts to handle different task characteristics, a method known as \textit{multi-LoRA architecture} (MoELoRA). Specifically, this method involves adding multiple LoRA modules as experts within the Transformer sub-layers~\cite{gao2024higherlayersneedlora}, and selectively activating weights based on input through a gating mechanism (Router), thereby enhancing performance of multi-task learning. Currently, multi-LoRA architecture~\cite{dou2023loramoe,luo2024moelora,li2024mixloraenhancinglargelanguage} have effectively captured and integrated multi-domain knowledge from multidimensional task scenarios, leading to performance improvements in complex downstream applications.

Despite its promise, MoELoRA may not effectively adapt the multi-task scenario, which can be distilled into two principal aspects: \textbf{(i) Training and Inference Efficiency.} Our observations show that MoELoRA fails to effectively balance performance against computational costs, contradicting the efficient characterization of PEFT, as illustrated in Figure \ref{fig:Overall of PEFT Methods Comparison} (training time is nearly \textbf{62\%} slower compared to LoRA). Additionally, multiplying the number of LoRA experts means introducing a proportional increase in matrix operations, which escalates training costs and inference latency. 
\textbf{(ii) Effectiveness of Expert Combination.} While advanced multi-LoRA architecture-based PEFT methods focus on adaptively selecting a subset of experts for updating, qualitative analysis~\cite{zuo2021taming} reveals that commonly-adopted mechanisms suffer from the notorious \textit{load imbalance} and \textit{overconfidence}. Gating mechanisms may not effectively learn task patterns and could lead to weight collapse, causing some experts to consistently dominate. Moreover, identical expert structures can lead to redundancy, raising concerns about the effectiveness of expert knowledge integration and transfer in MoELoRA across multidimensional task scenarios. Summing up, these limitations necessitate a reevaluation of MoELoRA and its solutions for handling multidimensional tasks, with the objective of achieving the right balance between effectiveness and efficiency.
% Current routers indicate the preferences of experts for tokens, highlighting a competitive dynamic among experts. However, routers often encounter challenges like random selection or overconfidence~\cite{zuo2021taming}, leading to noise-like behaviors, such as arbitrary assignments or concentrating information heavily on a few experts, contrary to the router's intended purpose. Analyses have revealed that solely depending on numerical comparisons to define competitive relationships among experts, without considering their intricate interactions, is inadequate for ensuring the effective transfer of expert knowledge to specific domains.
% In practical applications, MoE-LoRA faces the following primary challenges:
% (i) Identical structures tend to learn similar representations, limiting the knowledge density of the combined LoRA experts and hindering knowledge transfer and sharing among them.
% (ii) Multiplying the number of LoRA experts means introducing a proportional increase in matrix operations, which adds to training costs and inference latency.
% (iii) The complex interactions between experts are overlooked, failing to ensure the effective transfer of specific expert knowledge to corresponding domains.

To alleviate the aforementioned limitations, we propose a unified framework for efficient and effective multi-task learning, namely \includegraphics[height=0.12in, width=0.12in]{fig/cooperation.png} \textbf{\ourmethod{}}. Our bootstrapping philosophy involves treating the multiple experts as a ``\textit{Team}'', where through internal collaboration and competition among experts, we aim to enhance both efficiency and effectiveness respectively.  \ourmethod{} comprises two key components: \textbf{Efficient Collaboration Module}, which builds on the idea that the hierarchical relationship between the $A$ and $B$ matrices implies diversity in feature expression~\cite{hayou2024lora+}. We propose an asymmetric architecture for knowledge sharing and organization among experts.
Specifically, we treat $A$ as a domain-agnostic network with general knowledge and $B$ as a domain-specific network with unique task knowledge.
Matrix $A$ captures homogeneous features across tasks, playing a key role in learning and transmitting general knowledge, whereas Matrix $B$ concentrates on task-specific features, showcasing its expertise and efficient learning capacity within particular domains.  This allows different $B$ matrices to provide specialized supplements to $A$, enabling a ``plug-in'' based knowledge organization for collaborative experts. Compared to MoELoRA, this asymmetric knowledge expression strategy enhances training and inference efficiency through fewer matrix calculations; \textbf{Effective Competition Module}: Inspired by game theory~\cite{shapley1953value}, we introduce a competitive interaction mechanism to boost expert participation based on diverse task-aware inputs, addressing the shortcomings of overconfident routing in MoE. We employ the concept of Shapley values to foster competition among experts through finer-grained interactions, encouraging the effective transfer of domain-specific knowledge to corresponding downstream tasks.
By integrating both collaboration and competition, we ensure that internal experts work together as a ``\textit{Team}'', thus concurrently facilitating efficiency and effectiveness. 

To validate the effectiveness of \ourmethod{} in multi-task learning, we developed a \texttt{comprehensive multi} \texttt{-task evaluation} (CME) benchmark containing $2.5$ million samples, covering various domains and task types. 
In addition to single-modal fine-tuning, we also explored the feasibility of \ourmethod{} for visual instruction tuning on LLaVA-1.5~\cite{liu2024improved}. The experiments confirm that \ourmethod{} outperforms standard MoE-LoRA, providing the right balance between effectiveness and efficiency. Our contributions are as follows:
\begin{itemize}
    \item We designed a collaborative mechanism that facilitates ``plug-in'' knowledge organization and sharing, reducing computational costs.
    % cross-task knowledge sharing through shared general and expert modules while significantly reducing computational costs.
    \item We proposed a competition mechanism that adaptively adjusts the level of expert participation, emphasizing the effective transfer of knowledge to specific domains.
    % domain-specific knowledge to downstream tasks.
    \item We integrated a CME benchmark that encompasses multiple task types to evaluate PEFT methods.
    \item By integrating both collaborative and competitive mechanisms, \ourmethod{} enhances performance and alleviates efficiency bottlenecks in the multi-LoRA architecture.
\end{itemize}

\section{Related Work}

% \subsection{Large language models}
% Large language models (LLMs) \cite{2402.06196} nowadays excel in handling long sequences in natural language processing (NLP) tasks.
% In using transformer \cite{vaswani2017attention} architecture as the backbone, LLMs have made significant progress in tasks such as machine translation, setting new benchmarks for performance and training efficiency.
% Recently, apart from using transformer architecture, LLMs are gradually adopting other different architecture, such as Mamba Block \cite{mamba} and Test-Time Training (TTT) layers \cite{2407.04620}.
% One of the characteristics of LLMs is the massive amount of their parameters. As the size of LLMs scales up, finetuning them becomes a computationally expensive, time-consuming, and memory intensive process.
% Addressing the challenges posed by large model size has led to the investigation of various techniques, including parameter-efficient fine-tuning (PEFT) \cite{peft}, distillation, quantization, pruning, etc. We basically implement a PEFT-based method in our work.
\begin{figure*}[t]
    \centering
    \includegraphics[width=\linewidth]{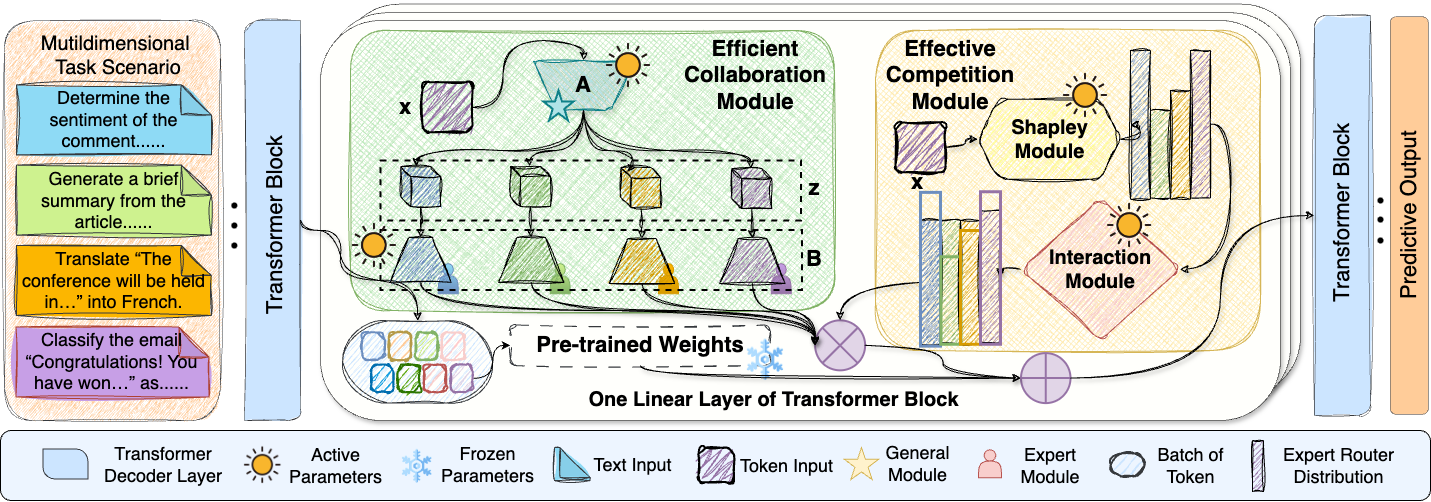}
    \caption{The architecture of \ourmethod{}. \ourmethod{} employs an asymmetric structure consisting of a general module and multiple expert modules as lightweight auxiliary modules to the pre-trained weights and enhances interactions between experts using a competition mechanism, enhancing the capability for multi-task learning.}
    \label{fig:fig3}
\end{figure*}

\subsubsection{Mixture-of-Experts.}
MoE integrates the outputs of multiple sub-models (experts) using a token-based routing mechanism~\cite{jacobs1991adaptive}. Shazeer et al.~\cite{shazeer2017outrageously,fedus2022switch} introduced a sparsely-gated top-k mechanism where the router activates a subset of experts for each input token, significantly reducing resource consumption during both training and inference. To balance expert loads, GShard~\cite{lepikhin2020gshard} and OpenMoE~\cite{xue2024openmoe} introduced importance and load losses to ensure fair load distribution among experts, reducing issues such as tail dropping and early routing learning. Additionally, the router's z-loss has been used to enhance training stability~\cite{zoph2022st}, and it addresses the expert balancing issue in multi-task models by maximizing mutual information between tasks and experts~\cite{chen2023mod}. Beyond token selection gating, Expert-Choice Gating allows experts to actively select the top-k tokens they will process, evenly distributing the load and avoiding the need for auxiliary losses~\cite{zhou2022mixture}. Recently, MoE has further explored potential in terms of the number of experts~\cite{he2024mixturemillionexperts} and multimodal fusion~\cite{lin2024moe}, becoming a focus of research.

\subsubsection{Parameter-Efficient Fine-Tuning.}
PEFT~\cite{he2021towards} reduces the dependency of fine-tuning Large Language Models (LLMs) on computational costs by introducing additional modules to replace updates to the large-scale pre-trained weights. Adapters~\cite{houlsby2019parameter} introduce extra feature transformations between blocks, prefix tuning~\cite{li2021prefix, liu2021p} updates parameters through prefixed learnable embeddings, and operations on pre-trained weights~\cite{liu2022few} also provide a feasible solution. Low-Rank Adaptation (LoRA)~\cite{hu2022lora} and its variants~\cite{yeh2024navigatingtexttoimagecustomizationlycoris,wu2024mixture} offer exceptional performance through low-rank matrix decomposition, and AdaLoRA~\cite{zhang2023adalora} seeks further optimization of embedding dimensions.

\subsubsection{Multi-LoRA Architectures.} Multi-LoRA architectures have also garnered widespread attention. Methods based on categorical assignments~\cite{zhao2024loraretriever,feng2024mixture,wu2024mixture} train multiple dedicated LoRAs that dynamically combine when handling complex tasks, providing robust performance. For general scenarios, researchers aim to introduce the dynamic capabilities of MoE, adaptively learning and combining multiple domain experts~\cite{luo2024moelora,tian2024hydralora,gao2024higherlayersneedlora}. In this work, we propose \ourmethod{}, designed to mitigate the efficiency limitations of Multi-LoRA architectures, offering enhanced performance and faster response times.

% \subsection{Shapley values}
% %夏普利值在深度学习的应用
% In the context of MoE, Shapley values~\cite{shapley1953value} provide a viable perspective for selecting experts and assessing their contributions, and are currently widely used to explain deep learning models. The SHAP method~\cite{lundberg2017unified} utilizes Shapley values to explain predictions made by machine learning models. Additionally, the study~\cite{sundararajan2020many} further explores the diverse applications of Shapley values in model explanation. Concurrently, Li et al.~\shortcite{li2022fine} demonstrates the feasibility of precisely adjusting the contributions and interactions of experts within MoE through multimodal, fine-grained semantic alignment. Furthermore, Gallardo et al.~\shortcite{gallardo2015games} illustrates the application of fuzzy set theory to Shapley values, adding flexibility and interpretability to dynamic decision-making in MoE routers. We utilize Shapley values to represent the marginal contributions of players, and through an interaction matrix with other players, we determine more appropriate participation weightings.

\section{Methods}
This section demonstrates the details of \ourmethod{}. Figure \ref{fig:fig3} illustrates the architecture of \ourmethod{}.

\subsection{Problem Formulation}
In a multi-task learning cenarios, Parameter-Efficient Fine-Tuning (PEFT) adapts to various application through a lightweight auxiliary module that is shared among tasks. This multi-task PEFT approach allows the model to remain compact while addressing multiple task requirements. Specifically, PEFT organizes shared auxiliary modules \(C_\text{aux}\) to a pre-trained layer \(C_\text{pre}\) for various types of tasks. The input sequence \(\bm{x} = [x_1, x_2, \ldots, x_N]\) is processed by the pre-trained layer and auxiliary module as follows:
\begin{equation}
    C_\text{mix}(\bm{x}; \theta_\text{pre}, \theta_\text{aux}) = C_\text{pre}(\bm{x}; \theta_\text{pre}) \oplus C_\text{aux}(\bm{x}; \theta_\text{aux})\,,
\end{equation}
where \(\theta_\text{pre}\) and \(\theta_\text{aux}\) denote the parameters of the pre-trained layer and the auxiliary module, respectively. \(\oplus\) represents combination strategies based on the method being used, which can be addition, multiplication, or concatenation.

During training, only the parameters of the auxiliary module are updated. This parameter update strategy maintains knowledge stability and reduces computational overhead:
\begin{equation}
    \theta_{\text{pre}} \leftarrow \theta_{\text{pre}},\;\theta_{\text{aux}} \leftarrow \theta_{\text{aux}} - \eta \nabla_{\theta_{\text{aux}}}\mathcal{L}(\bm{y}, \bm{y}_{\text{gt}})\,,
\end{equation}
where \( \eta \) represents the learning rate and target optimization function \( \mathcal{L} \) assesses the deviation between the predicted output \( \bm{y} \) and the ground truth \( \bm{y}_{\text{gt}} \).

\subsection{Preliminaries}
\subsubsection{Low-Rank Adaptation.} 
LoRA~\cite{hu2022lora} captures downstream data features by introducing a pair of low-rank matrices as auxiliary modules for the pre-trained weights. The core idea of LoRA is to decompose the auxiliary weight matrix $\Delta W\in\mathbb{R}^{d_\textbf{in}\times d_\textbf{out}}$ of the linear layer into two matrices, $A\in\mathbb{R}^{d_\textbf{in}\times r}$ and $B\in\mathbb{R}^{r\times d_\textbf{out}}$ with r $\ll \min\{d_\textbf{in}, d_\textbf{out}\}$, reducing the number of learnable parameters. Assuming the origin input to pre-trained weights is $\bm{x}\in\mathbb{R}^{N\times d_\textbf{in}}$ and the output $\bm{h}\in\mathbb{R}^{N\times d_\textbf{out}}$ with LoRA can be represented as:
\begin{equation}
    \bm{h} = \bm{x}W_0 + \bm{x}\Delta W = \bm{x}W_0 + \bm{x}AB\,,
\end{equation}
where matrix $A$ is initialized with a random Gaussian distribution and matrix $B$ as a zero matrix to ensure that LoRA does not affect the original output at the start of training. Typically, \(\Delta W\) is scaled by \(\alpha/r\), using a scaling factor $\alpha$ to adjust the impact of the LoRA module.

% According to block matrix multiplication, A can be decomposed into a series of column vectors $[A_1, A_2, \ldots, A_r]$ and $B$ into a series of row vectors $[B_1^T, B_2^T, \ldots, B_r^T]^T$. The output $h$ can be re-expressed as a sum of submatrix multiplications:
% \begin{equation}
%     h = xW_0 + \sum_{i=1}^r \omega_ixA_iB_i,
% \end{equation}
% where $\omega_i = 1$. Viewing $A_i$ and $B_i$ as a single expert indicates that LoRA is a specialized type of MoE utilizing a static router that treats the importance of all experts equally.. Thus, LoRA can be regarded as a framework of absolute cooperation within the MoE.

\subsubsection{Mixture of Experts.}
MoE~\cite{fedus2022switch} greatly expands the model scale while activating only a small number of parameters. In large models (LMs), MoE duplicates the Feed-Forward Network (FFN) to create a collection of experts, facilitating the transfer of specific knowledge to downstream tasks, thereby enhancing model performance without significantly increasing training time and inference latency. Specifically, MoE constructs a set of \(k\) experts, \(\{E_i\}_{i=1}^k\), and utilizes a router \(R\) with Softmax normalization to dynamically allocate a set of weights \(\bm{\omega}\) for token participation:
\begin{equation}
    \omega_i = \frac{e^{(R_i(\bm{x}; \theta_R))}}{\sum_{j=1}^k e^{(R_j(\bm{x}; \theta_R))}}\,,
\end{equation}
where \(\theta_R\) represents the parameters of the router, which is typically a fully connected layer.
The output of the FFN layer can be represented as \( \bm{y} = C_{\text{ffn}}(\bm{x}; \theta_{\text{ffn}}) \). 

Correspondingly, the output with MoE is as follows:
\begin{equation}
    \bm{y} = C_{\text{MoE}}(\bm{x};\theta_R,\{\theta_\text{ffn}^i\}_{i=1}^k) = \sum_{i=1}^k \omega_i E_i(\bm{x}; \theta^i_{\text{ffn}})\,,
\end{equation}
where $E_i$ represents $i$-th extended FFN expert, and \(\theta^i_{\text{ffn}}\) denotes the parameters of the \(i\)-th expert.

\subsection{\includegraphics[height=0.12in, width=0.12in]{fig/cooperation.png} \textbf{\ourmethod{}}}
\ourmethod{} facilitates efficient collaboration and effective competition among experts, optimizing the mechanisms for knowledge sharing and transfer to boost performance:
\begin{equation}
    C_{\text{mix}}(\bm{x}; W_0, \theta_\text{col},\theta_\text{cop}) = \bm{x}W_0 + C_{\text{aux}}(\bm{x}; \theta_\text{col},\theta_\text{cop})\,,
\end{equation}
where $\theta_\text{col}$ represents parameters of efficient collaboration module $\mathcal{M}_\text{col}$ and $\theta_\text{cop}$ represents parameters of effective competition module $\mathcal{M}_\text{cop}$.

\subsubsection{Efficient Collaboration among Experts.}
We first analyze MoELoRA, which adopts an adaptive collaboration approach, dynamically combining LoRA expert knowledge \(\{E\}_{i=1}^k\) through a router mechanism. The combined knowledge is added as a bypass to the pretrained weights. Specifically, MoELoRA constructs multiple identical expert pairs \(\{A_i, B_i\}_{i=1}^k\) to perform multi-task learning and the mechanism of MoELoRA is illustrated as follows:
\begin{equation}
    C_{\text{aux}}(\bm{x}; \theta_R,\{A_i,B_i\}_{i=1}^k) = \sum_{i=1}^k \omega_i E_i(\bm{x}; A_i, B_i)\,,
\end{equation}
where \(E_i(\bm{x}; A_i, B_i) = \bm{x}A_iB_i\), and \(\bm{\omega}\) represents the normalized output of the router adaptively learned from tasks.

In fact, regarding MoELoRA, we have two key observations: (i) Based on the stacking of multiple LoRA experts, MoELoRA introduces an additional approximately 2$*k$ matrix operations, significantly impairing the GPU's parallel processing capabilities. For example, in our CMT benchmark, $k$ values of 4 or greater are nearly impossible to train (when $k$ equals 2, 4, and 8, MoELoRA introduced additional training times of \textbf{19\%, 62\%, and 138\%}, respectively, compared to LoRA). (ii) The independence among experts leads to learning redundant knowledge, evidenced by achieving \textbf{98.5\%} performance on the CMT benchmark as Table 1, when only the most advantageous experts (Top-1) are retained, which dilutes the collective expressive power of the expert ensemble. These scenarios prevent MoELoRA from effectively balancing between efficiency and performance.

Considering the structural hierarchy between $A$ and $B$, \ourmethod{} designs a collaboration module aimed at facilitating hierarchical collaboration between them. The general module (matrix A) captures homogeneous features across tasks, responsible for learning and transmitting domain-agnostic general knowledge; the expert modules (matrix B) considered as domain-specific plugins capture and promote corresponding knowledge transfer in specialized domains.

\ourmethod{} defines matrix $A \in \mathbb{R}^{d_\text{in} \times r_A}$ and $k$ matrices $B_i \in \mathbb{R}^{r_B \times d_\text{out}}$, where $r_A = kr_B$. The input $\bm{x}$ is processed through matrix $A$ to compute an intermediate state $\bm{z}=xA$, where $\bm{z} \in \mathbb{R}^{N \times r_A}$. Then $\bm{z}$ is evenly split into $k$ segments along its last dimension, a process we refer to as ``\textit{split}'':
\begin{equation}
    z_i =\text{\textit{split}}(\bm{z})_i= \bm{z}_{(i-1)r_B + 1 : ir_B}.
\end{equation}

Subsequently, each segment $z_i$ undergoes a linear transformation through its corresponding matrix $B_i$. The final partial output $h_i\in\mathbb{R}^{N \times d_\text{out}}$ as below:
\begin{equation}
    h_i = \mathcal{M}_\text{col}(\bm{x};A,B_i) = \text{\textit{split}}(\bm{x}A)_iB_i.
\end{equation}
Assuming expert weights is $\bm{\omega}$, the final output of the collaboration module can be represented as $\bm{h}=\sum_{i=1}^k\omega_ih_i$.
Such an operation is considered a knowledge organization and forward transfer by the ``\textit{Team}''.

\begin{table}[t]
\centering
% \captionsetup{font={small,stretch=1}, labelfont={bf}}
% \caption{Abalation study. }
\resizebox{0.99\columnwidth}{!}{% adjust width
\renewcommand{\arraystretch}{1.25}
    \begin{tabular}{c|cccc|c||c}
        \Xhline{1.5pt}
        % \rowcolor[HTML]{DAE0FB}
         \textbf{Expert ID}&\textbf{1}&\textbf{2}&\textbf{3}&\textbf{4}&\textbf{Top-1}&\textbf{All}\\
        \hline
        \hline
         \textbf{Performance} & 41.69 & 47.14 & 44.37 & 39.83 & \underline{58.78} & \textbf{59.96}\\
        
        % Rank&
        % $\mathcal{M}_\textbf{p}${\ding{55}}$\mathcal{M}_\textbf{l}${\ding{55}}&
        % $\mathcal{M}_\textbf{p}${\ding{51}}$\mathcal{M}_\textbf{l}${\ding{55}}&
        % $\mathcal{M}_\textbf{p}${\ding{55}}$\mathcal{M}_\textbf{l}${\ding{51}}&
        % $\mathcal{M}_\textbf{p}${\ding{51}}$\mathcal{M}_\textbf{l}${\ding{51}}\\
        % \hline
        % \hline

        % 8&57.65&58.18&\underline{58.27}&\textbf{58.31}\\
        % 16&59.08&59.25&\underline{59.77}&\textbf{59.95}\\
        % 32&59.69&59.77&\underline{60.24}&\textbf{60.29}\\
        % 64&58.88&\textbf{59.07}&58.87&\underline{58.94}\\
        \Xhline{1.5pt}
    \end{tabular}
    }
    \caption{Expert redundancy analysis of MoELoRA.}
    \label{tab:tab4}
\end{table}

% Collaboration module allows each expert to concentrate deeply on the unique characteristics of diverse domains, thereby improving the model's understanding of specific tasks. Initially processing the input \(x\) through a shared general matrix \(A\) enables the model to capture and leverage common knowledge across different tasks, thus maintaining its generalization capabilities. Furthermore, by evenly partitioning the intermediate state \(z\) and processing it through respective \(B_i\) matrices, the model facilitates dynamic knowledge transfer across different domains while preserving overall parameter efficiency. The collaboration module enables finer-grained optimization by coordinating the adjustments of the experts, allowing for dynamic adaptation to actual needs.

Unlike the fully symmetric structure of MoELoRA, the efficient collaborative module allows general modules and expert modules to adaptively organize team knowledge to cope with multi-task scenarios. The general module captures domain-independent common knowledge and maintains generalization performance in complex scenarios. Subsequently, the expert modules provide specialized knowledge supplementation and organization based on "plug-in" action, effectively capturing and integrating task-specific details, thereby improving the efficiency of knowledge transfer. Additionally, this collaborative module significantly reduces computational costs by decreasing matrix operations, requiring only \textbf{87\%}, \textbf{70\%}, and \textbf{63\%} of the training time of MoELoRA with the same number of experts when \(k\) is 2, 4, and 8 respectively, achieving the efficiency objective.
\begin{table*}
    \centering
    \setlength{\tabcolsep}{1mm}
    \fontsize{9}{11}\selectfont
    % \begin{tabular}{c||c|c|c||c|c|c|c|c|c|c|c|c|c|c|c||c}
    \resizebox{2.1\columnwidth}{!}{
    \begin{tabular}{c||cccc||ccccccccccc||c}
         \Xhline{1.5pt}
        % \multirow{2}{*}{Method} & \makecell{Text\\Summarization} & \multicolumn{2}{c|}{\makecell{Sentiment\\Classification}} & \multicolumn{3}{c|}{NLI} & \makecell{Coreference\\Resolution} &  \multicolumn{4}{c|}{Closed Book QA} & \multirow{2}{*}{Avg.} \\
        % \cline{2-12}
        %  & OAI-Sum & Emo & IMDB & ANLI & QQP & RTE & WinG & ARC & WQA & NQ & TQA & \\ 
        % \rowcolor[HTML]{DAE0FB}
        \textbf{Method}& \textbf{MoE} & \textbf{Rank} & \textbf{Time} & \textbf{Params\%} & \textbf{OAI-Sum} & \textbf{IMDB} & \textbf{ANLI} & \textbf{QQP} & \textbf{RTE} & \textbf{WinG} & \textbf{ARC} & \textbf{WQA} & \textbf{NQ} & \textbf{TQA} & \textbf{MMLU} & \textbf{Avg.} \\ 
        \hline
        \hline
        % \rowcolor[HTML]{E0E0E0}
        % Full Fine-Tuning & 28.6 & 90.1 & 96.5 & 60.5 & 88.3 & 88.7 & 74.5 & 72.7 & 52.1 & 26.5 & 39.2& - & 65.25 \\ 
        % \cellcolor[HTML]{E0E0E0}
        Prompt-Tuning& {\Large \ding{55}}& -  & 23h & 0.02 & 25.3 & 91.1 & 44.2 & 77.0 & 65.4 & 59.7 & 54.8 & 38.7 & 16.2 & 19.4 & 31.2&47.55 \\ 
        % \cellcolor[HTML]{E0E0E0}
        IA3 &  {\Large \ding{55}} & - & 24h & 0.03 & 26.4 & 92.0 & 48.7 & 78.3 & 68.1 & 61.5 & 55.1 & 37.7 & 18.8 & 19.5 &34.9& 49.18 \\

        % \cellcolor[HTML]{E0E0E0}
        % LoRA* & 27.1 & 95.4 & 58.1 & 87.9 & 88.3 & 72.0 & 70.1 & 40.4 & 21.0 & 29.0 &40.4& 57.25&32&1.15&31h  \\ 
        \hdashline
        
        LoRA&{\Large \ding{55}}&32 & 25h& 0.67& 27.2 & 95.6 & 57.6 & 84.9 & 87.0 & 65.8 & 68.2 & 47.1 & 23.3 & 34.7&40.4 & 57.44\\
        LoRA&{\Large \ding{55}}&128 & 26h&2.68& 27.3 & 95.6 & 56.8 & \underline{87.4} & 85.7 & 71.6 & 70.8 & 47.2 & 25.2 & 36.8&42.5 & 58.81\\
        % \cellcolor[HTML]{E0E0E0}
        AdaLoRA&  {\Large \ding{55}}& 128 &30h & 2.56 & \underline{27.4} & 95.5 & 57.2 & 87.0 & 86.3 & 72.1 & 71.1 & 46.8 & {25.5} & 35.2&\underline{42.9}& 58.82\\ 
        MoSLoRA& {\Large \ding{55}} & 128 & 28h&2.70& 27.3 & {95.6} & 58.3 & 86.8 & 86.6 & \underline{73.2} & 71.9 & 47.4 & 25.8 & 38.4&41.4 & 59.34 \\ 
        HydraLoRA &{\Large \ding{51}} & 32& 34h&1.84 & \textbf{27.6} & \textbf{95.9} & 57.8 & 86.5 & \textbf{87.2} & 70.1 & 70.2 & {50.6} & 24.6 & 37.0 &42.2& 59.06 \\

        MoELoRA & {\Large \ding{51}}&32&42h&2.71 & \underline{27.4} & 95.5 & \textbf{59.3} & 87.2 & 86.1 & {72.9} & {71.8} & 50.1 & 25.1 & \underline{38.4} &42.8& {59.69}  \\ 
        \hdashline
        % \cellcolor[HTML]{D9EAD3}
        \textbf{\ourmethod{}}&{\Large \ding{51}}&16&28h&1.35 & \underline{27.4} & \textbf{95.9} & \underline{59.2} & {86.6} & {87.0} & \textbf{73.1} & \underline{73.1} & \underline{51.3} & \underline{25.9} & {37.1}&{42.8}  & \underline{59.95} \\ 
        % \cellcolor[HTML]{D9EAD3}
        \textbf{\ourmethod{}}&{\Large \ding{51}}&32&29h&2.71 & \textbf{27.6} & \underline{95.7} & {58.9} & \textbf{87.5} & \underline{87.1} & \textbf{73.8} & \underline{72.3} & \textbf{51.8} & \textbf{26.4} & \textbf{38.8}&\textbf{43.3}  & \textbf{60.29} \\ 
        % \hline \hline

        % LoRA& 27.3 & 95.6 & 56.8 & 87.4 & 85.7 & 71.6 & 70.8 & 47.2 & 25.2 & 36.8&42.5 & 58.81&128&2.68&26h \\ 
        % MoSLoRA & 27.4 & 95.6 & 58.3 & 86.8 & 86.6 & 73.2 & 71.9 & 47.4 & 26.4 & 38.4&41.4 & 59.40&128&2.70&27h \\ 
        % HydraLoRA & 27.1 & 95.6 & 57.9 & 88.2 & 87.4 & 72.7 & 73.1 & 50.3 & 25.8 & 37.3 &42.9& 59.85&64&3.68&36h \\ 
        % MoELoRA & 27.4 & 95.5 & 60.3 & 87.2 & 86.1 & 72.9 & 71.8 & 50.1 & 25.1 & 38.4 &42.8& 59.80&32&2.71&42h \\ 
        % % \hline
        % % \cellcolor[HTML]{D9EAD3}
        % % LoRAMoG & \textbf{28.0} & \textbf{96.0} & 58.3 & 87.0 & \textbf{88.2} & 72.2 & \textbf{72.2} & 50.6 & 26.1 & 38.5 &42.8& 63.93\\ 
        % \cellcolor[HTML]{D9EAD3}
        % TeamLoRA & 27.6 & 95.7 & 58.9 & 87.5 & 87.0 & 73.8 & 72.3 & 51.8 & 26.4 & 38.8&43.3  & \textbf{60.28}&32&2.71&29h \\ 
        \Xhline{1.5pt}
    \end{tabular}}
    \caption{Performance comparison of TeamLoRA and other PEFT methods on the CME benchmark. \textit{MoE} indicates whether the MoE architecture is used, \textit{Rank} represents the dimension of the expert modules ($r_B$ for \ourmethod{} and $r$ for other methods), \textit{Time} denotes the training time of the model on 8$\times$A800 GPUs, and \textit{Params\%} represents the number of learnable parameters. The best results are marked in bold, while the second-best results are underlined.}
    \label{tab:tab1}
\end{table*}
\subsubsection{Effective Competition among Experts.}

\begin{table*}[h!]
    \centering
    \setlength{\tabcolsep}{1mm}
    \fontsize{9}{11}\selectfont
    \resizebox{2.1\columnwidth}{!}{
    \begin{tabular}{c||c|cccccc|||c||c|cccccc}
        \Xhline{1.5pt}
        % \rowcolor[HTML]{DAE0FB}
        \textbf{Rank} & \textbf{Method} & \textbf{OAI-Sum} & \textbf{IMDB} & \textbf{QQP} & \textbf{WinG} & \textbf{NQ} &\textbf{TQA}&\textbf{Rank} & \textbf{Method} & \textbf{OAI-Sum} & \textbf{IMDB} & \textbf{QQP} & \textbf{WinG} & \textbf{NQ}& \textbf{TQA}\\ 
        \hline
        \hline
        32
        & LoRA & 27.2 & \underline{95.6} & 84.9 & 65.8 & \underline{23.3} &\underline{34.7}&
        64
        & LoRA & \textbf{27.4} & \underline{95.7} & \underline{86.4} & \underline{70.2} & \underline{25.6} & 35.5 \\ 
        \hdashline
        \multirow{2}{*}{8}& MoELoRA & \underline{27.3} & 95.5 & \textbf{86.3} & \underline{67.8} & 21.9&33.7 & \multirow{2}{*}{16}& MoELoRA & 27.7 & 95.6 & 86.3 & 69.5 & 24.3 &\underline{36.4} \\ 
         & 
         % \cellcolor[HTML]{D9EAD3}
         \textbf{\ourmethod{}} & \textbf{27.9} & \textbf{96.1} & \textbf{86.3} & \textbf{68.7} & \textbf{24.0}&\textbf{35.7} &  &
         % \cellcolor[HTML]{D9EAD3} 
         \textbf{\ourmethod{}} & \textbf{27.4} & \textbf{95.9} & \textbf{86.6} & \textbf{73.1} & \textbf{25.9} & \textbf{37.1} \\ 
        \hline
        128
        &LoRA & 27.3 & \underline{95.6} & \underline{87.4} & 71.6&\underline{25.2} & 36.8 &
        256
        & LoRA & 26.3 & \underline{96.0} & \underline{87.8} & 71.7 & 17.5 & 23.8 \\ 
        \hdashline
        \multirow{2}{*}{32}& MoELoRA & \underline{27.4} & 95.5 & 87.2 & \underline{72.9} & 25.1 & \underline{38.4} &\multirow{2}{*}{64} & MoELoRA & \textbf{26.9} & \textbf{96.2} & 87.3 & \underline{71.8}& \underline{21.8} & \underline{35.1}\\ 
         & 
         % \cellcolor[HTML]{D9EAD3}
         \textbf{\ourmethod{}} & \textbf{27.6} & \textbf{95.7} & \textbf{87.5} & \textbf{73.8} & \textbf{26.4} & \textbf{38.8} &  & 
         % \cellcolor[HTML]{D9EAD3}
         \textbf{\ourmethod{}} & \textbf{26.9} & 95.4 & \textbf{88.1} & \textbf{71.9} & \textbf{21.9} & \textbf{35.5} \\ 
       \Xhline{1.5pt}
    \end{tabular}}
    \caption{Performance of different methods across various tasks with different ranks.}
    \label{tab:tab2}
\end{table*}

% Current routers indicate the preferences of experts for tokens, highlighting a competitive dynamic among experts. However, routers often encounter challenges like random selection or overconfidence~\cite{zuo2021taming}, leading to noise-like behaviors, such as arbitrary assignments or concentrating information heavily on a few experts, contrary to the router's intended purpose. Analyses have revealed that solely depending on numerical comparisons to define competitive relationships among experts, without considering their intricate interactions, is inadequate for ensuring the effective transfer of expert knowledge to specific domains. Therefore, we propose a game-theoretic interaction mechanism that dynamically adjusts the competitive outcomes for tokens among experts, based on their interactive relationships.

Common routing mechanisms have key flaws such as inefficiency in allocation and knowledge silos~\cite{zuo2021taming}, which contradict the design philosophy. To address this, we introduc a shapley-based mechanism~\cite{shapley1953value} that actively shapes expert competition based on adaptive interactions. This approach prevents centralized decision-making and promotes the effective transfer of expertise to specific downstream tasks. By dynamically adjusting input distribution and expert responsibilities, the competition module ensures more effective and equitable knowledge transfer across tasks.

We first introduce the concept of \textit{fuzzy Shapley values} to offer a perspective on how routers assess the marginal contributions of experts. Unlike the traditional binary participation (participation or absence), \textit{fuzzy Shapley values} permit participation degrees to range from \(0\) to \(1\). The following equation represents the marginal contribution of experts:
\begin{equation}
    \phi_i(\bm{x};\omega_i) = \int_{s} \left( v_i(\bm{x},w_i, s) - v_i(\bm{x},0, s) \right) \, ds\,,
\end{equation}
where \(\phi_i(\bm{x};\omega_i)\) represents the marginal contribution of expert \(i\) with participation degree \(\omega_i\), and \(s\) denotes the space of possible participation degrees for the remaining experts, satisfying \(\sum_js_j=1-\omega_i\) and \(j \neq i\). \(v_i(\bm{x},\omega_i, s)\) represents the total payoff from the combined participation $\{\omega_i\} + s$.

From the perspective of shapley values, the mechanism of the router can be understood as assessing the average marginal contributions of each expert across all possible combinations of experts. This provides a theoretical basis for the allocation of activation weights and highlights the importance of considering synergistic effects among experts. Although calculating shapley values is an NP-hard problem in practical applications, we can use an MLP as an approximation module for fuzzy Shapley values, estimating the marginal contributions of each expert:
\begin{equation}
    \phi_i(\bm{x};\theta_S) \leftarrow \text{Softmax}(S(\bm{x};\theta_S))_i\,,
\end{equation}
where $\phi_i$ represents the fuzzy Shapley value of the $i$-th expert and $S$ represents Shapley value calculator.

% \begin{equation}
% \psi_{T}(x) = \sum_{S \subseteq N \setminus T} \frac{|S|!(|N|-|S|-|T|)!}{|N|!} \left(\mu(S \cup T) - \sum_{J \subset T} (-1)^{|T|-|J|} \mu(S \cup J)\right),
% \end{equation}

To fully capture the competitive dynamics among experts, we introduce an interaction matrix that evaluates and adjusts their interactions. This matrix captures the mutual influences among experts and adjusts their participation based on Shapley interactions. Specifically, the interaction matrix $M$ is designed to adaptively adjust each expert's participation based on their competitive relationships, as detailed below:
\begin{equation}
    \omega_i = \mathcal{M}_\text{cop}(\bm{x};\theta_S,M)= \sum_{j=1}^{k} M_{ij} \phi_j(\bm{x};\theta_S)\,,
\end{equation}
where $\omega_i$ represents the adjusted optimal degree of participation, and $M_{ij}$ denotes the element in the interaction matrix reflecting the influence of expert $j$ on expert $i$. The interaction matrix \( M \) is initialized with a uniform distribution, with all diagonal elements set to $\bm{1}$ for baseline self-influence. $M$ is a learnable matrix that adapts during the training process to fully account for synergistic effects among experts and adequately captures the competitive relationships.

Ultimately, the output of \ourmethod{} is represented as:
\begin{equation}
    \bm{h} = \bm{x}W_0 + \mathcal{M}_\text{col}(\bm{x};A,\{B_i\})\odot \mathcal{M}_\text{cop}(\bm{x};\theta_S,M)\,,
\end{equation}
where $\odot$ represents the element-wise product.

\section{Experiments}

% \begin{table}
%     \centering
%     \setlength{\tabcolsep}{1mm}
%     \fontsize{9}{11}\selectfont
%     \begin{tabular}{c|ccc||c|ccc}
%         \hline
%         \multirow{1}{*}{Method} & \multirow{1}{*}{A} & \multirow{1}{*}{B} & \multirow{1}{*}{Gate}&\multirow{1}{*}{Method} & \multirow{1}{*}{A} & \multirow{1}{*}{B} & \multirow{1}{*}{Gate} \\
%         \hline
%         \hline
%         LoRA & 1 & 1 & I & MoSLoRA & 1 & 1& M \\
%         HydraLoRA & 1 & 4 & R & MoELoRA & 4 & 4 & R \\
%         \hline
%         \cellcolor[HTML]{D9EAD3}
%         TeamLoRA* & 4 & 4 & R  &\cellcolor[HTML]{D9EAD3} TeamLoRA&4&4&C\\
%         \hline
%     \end{tabular}

%     \caption{Comparison of different MoE-based methods highlighting performance metrics across various parameter ratio (the proportion of training parameters relative to the total model parameters), training time (the duration required to complete one training session on 8 A800 GPUs), number of $A$/$B$ and specialized gating mechanisms (I: Identity mapping; M: Mixture matrix; R: Router; S: Shapley interaction module).}
%     \label{tab:model detail}
% \end{table}

% We conduct rigorous experiments and analyses on the performance of \ourmethod{} in multi-task scenarios, verifying its significant advantages over other PEFT methods, particularly those based on Multi-LoRA Architectures. 

% Experimental results indicate that \ourmethod{} achieves a balance between efficiency and effectiveness, demonstrating strong potential in multi-task learning scenarios.

\subsection{Benchmark and Setting}
\noindent\textbf{Benchmark.} All PEFT methods used the 2.5M training set from 22 datasets effectively organized by CME(refer to Appendix A) and were comprehensively evaluated on tasks across 11 different tasks: OpenAI-Summarize-TLDR~\cite{stiennon2020learning}, IMDB~\cite{maas-EtAl:2011:ACL-HLT2011}, ANLI~\cite{nie2019adversarial}, QQP~\cite{wang2017bilateral}, RTE~\cite{wang2019gluemultitaskbenchmarkanalysis}, WinoGrande~\cite{sakaguchi2021winogrande}, ARC~\cite{allenai:arc}, WebQA~\cite{li2016datasetneuralrecurrentsequence}, NQ~\cite{kwiatkowski2019natural}, TriviaQA~\cite{2017arXivtriviaqa}, and MMLU~\cite{hendryckstest2021}.

\begin{table}[t]
\centering
% \captionsetup{font={small,stretch=1}, labelfont={bf}}
% \caption{Abalation study. }
\resizebox{0.98\columnwidth}{!}{% adjust width
\renewcommand{\arraystretch}{1.25}
    \begin{tabular}{c|c||c|c|c|c}
        \Xhline{1.5pt}
        % \rowcolor[HTML]{DAE0FB}
         \textbf{Cop} & \textbf{Col}& \textbf{Avg}$ _{r=8}$ & \textbf{Avg}$ _{r=16}$ &\textbf{Avg}$ _{r=32}$&\textbf{Avg}$ _{r=64}$\\
        \hline
        \hline
         -&-& 57.65  & 59.08 &  59.69 & 58.88\\
         % \cellcolor[HTML]{D9EAD3}
         \faCheckCircle & - & 58.18&59.25&59.77&\textbf{59.07}\\
         - & 
         % \cellcolor[HTML]{D9EAD3}
         \faCheckCircle &\underline{58.27}&\underline{59.77}&\underline{60.24}&58.87\\
         % \cellcolor[HTML]{D9EAD3}
         \faCheckCircle &
         % \cellcolor[HTML]{D9EAD3}
         \faCheckCircle &\textbf{58.31}&\textbf{59.95}&\textbf{60.29}&\underline{58.94}\\
        
        % Rank&
        % $\mathcal{M}_\textbf{p}${\ding{55}}$\mathcal{M}_\textbf{l}${\ding{55}}&
        % $\mathcal{M}_\textbf{p}${\ding{51}}$\mathcal{M}_\textbf{l}${\ding{55}}&
        % $\mathcal{M}_\textbf{p}${\ding{55}}$\mathcal{M}_\textbf{l}${\ding{51}}&
        % $\mathcal{M}_\textbf{p}${\ding{51}}$\mathcal{M}_\textbf{l}${\ding{51}}\\
        % \hline
        % \hline

        % 8&57.65&58.18&\underline{58.27}&\textbf{58.31}\\
        % 16&59.08&59.25&\underline{59.77}&\textbf{59.95}\\
        % 32&59.69&59.77&\underline{60.24}&\textbf{60.29}\\
        % 64&58.88&\textbf{59.07}&58.87&\underline{58.94}\\
        \Xhline{1.5pt}
    \end{tabular}
    }
    \caption{Ablation analysis for collaboration and competition modules.}
    \label{tab:tab4}
\end{table}

% \begin{table}[t]
% \centering
% \captionsetup{font={small,stretch=1}, labelfont={bf}}
% \caption{Abalation study. }
% \resizebox{1\columnwidth}{!}{% adjust width
% \renewcommand{\arraystretch}{1.25}
% \begin{tabular}{l||ccc||cccc}
% \Xhline{1.25pt}
% \multirow{2}{*}{\textbf{\Large{Method}}}  & 
% \multicolumn{3}{c||}{\texttt{Data Selection}} & \multicolumn{4}{c}{\texttt{Benchmarks}}  \\
% \Xcline{2-4}{0.25pt} \Xcline{5-8}{0.25pt}
% & \textbf{Human} & \textbf{Rewrite} & \textbf{Review} & \textbf{VQAv2} & \textbf{SQA } & \textbf{MME } & \textbf{MM-Vet}  \\
% \Xhline{1.25pt}
% Align$^2$LLaVA & \faCheckCircle  &                &                & - & - & - & -  \\
% Align$^2$LLaVA & \faCheckCircle  & \faCheckCircle &                & - & - & - & -  \\
% Align$^2$LLaVA & \faCheckCircle  & \faCheckCircle & \faCheckCircle & - & - & - & -  \\
% \Xhline{1.25pt}
% \end{tabular}%
% }

% \label{tab:method_comparison}
% \end{table}

\noindent\textbf{Training Details.} 
We selected the LLaMA-2 7B~\cite{touvron2023llama} as the base model and continued pre-training it on the expanded Chinese LLaMA-2-7B corpus~\cite{Chinese-LLaMA-Alpaca} to enhances the model's knowledge capacity and multilingual capability by expanding the vocabulary and incorporating general corpora. To ensure fairness, for all LoRA-based PEFT methods, we added parameters only to the FFN module and maintained nearly identical parameter increments within the same experimental setup to minimize the potential impact of parameter size on performance. All experiments were conducted on 8$\times$A800 GPUs, using the same hyperparameter(listed in Appendix B) settings.

% and averaging across multiple seed settings to ensure the model's generalization ability.

\noindent\textbf{Comparison of Methods.}
% \subsection{Comparison of Methods}
To evaluate the superiority of \ourmethod{}, we selected several prominent PEFT methods, including Prompt-Tuning~\cite{lester2021powerscaleparameterefficientprompt}, IA3~\cite{liu2022few}, LoRA~\cite{hu2022lora}, MoSLoRA~\cite{wu2024mixture} and AdaLoRA~\cite{zhang2023adalora}. We also primarily compared methods utilizing MoE mechanisms: MoELoRA(multi-lora architecture), HydraLoRA~\cite{tian2024hydralora}. It's worth noting that MoSLoRA provides insights similar to MoELoRA from the perspective of matrix decomposition. We further conducted evaluations on Llama-3 8B~\cite{dubey2024llama} and LLaVA-1.5 7B~\cite{liu2024improved} for further exploration.

\begin{figure}[t]
    \centering
    \includegraphics[width=0.98\linewidth]{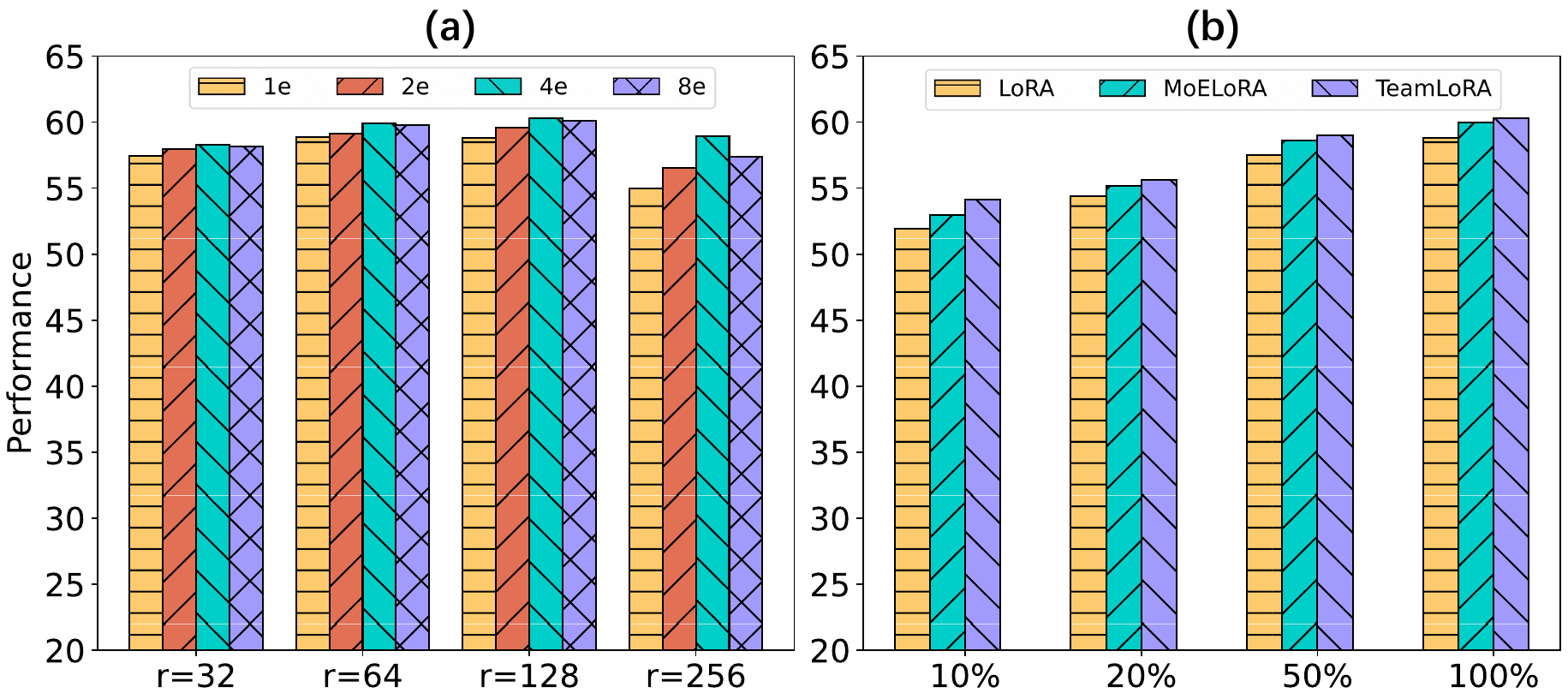}
    \caption{Stability analysis. (a) illustrates how the number of expert modules impact performance. (b) shows the performance comparison of \ourmethod{} under different data scales.}
    \label{fig:fig4}
\end{figure}

\begin{figure}[t]
    \centering
     \includegraphics[width=0.96\linewidth]{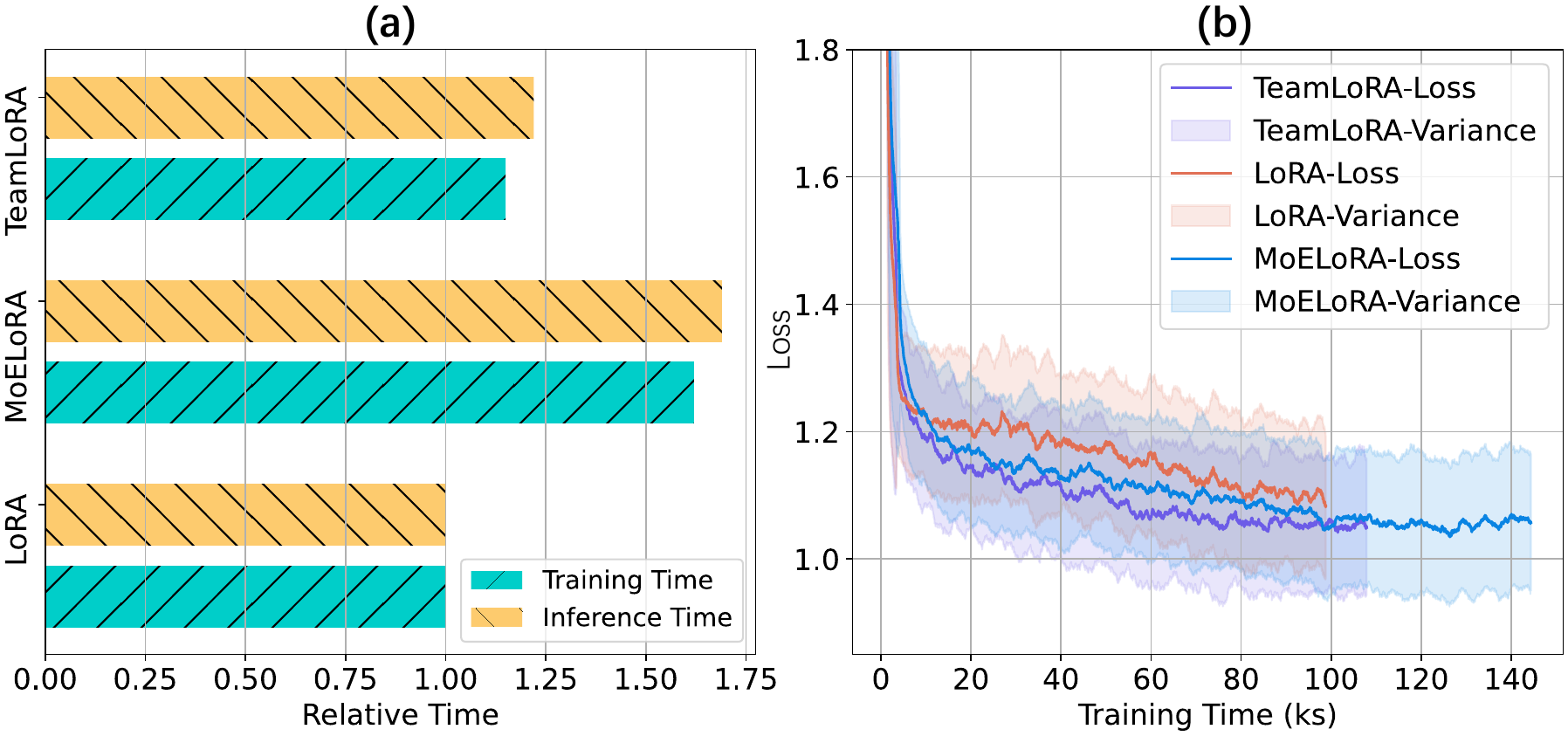}
    \caption{Visualization of Efficiency and Loss. (a) describes the relative training and inference latency of \ourmethod{} and MoELoRA compared to LoRA. (b) displays the loss convergence.}
    \label{fig:fig4}
\end{figure}

\begin{figure*}
    \centering
    \includegraphics[width=0.58\linewidth]{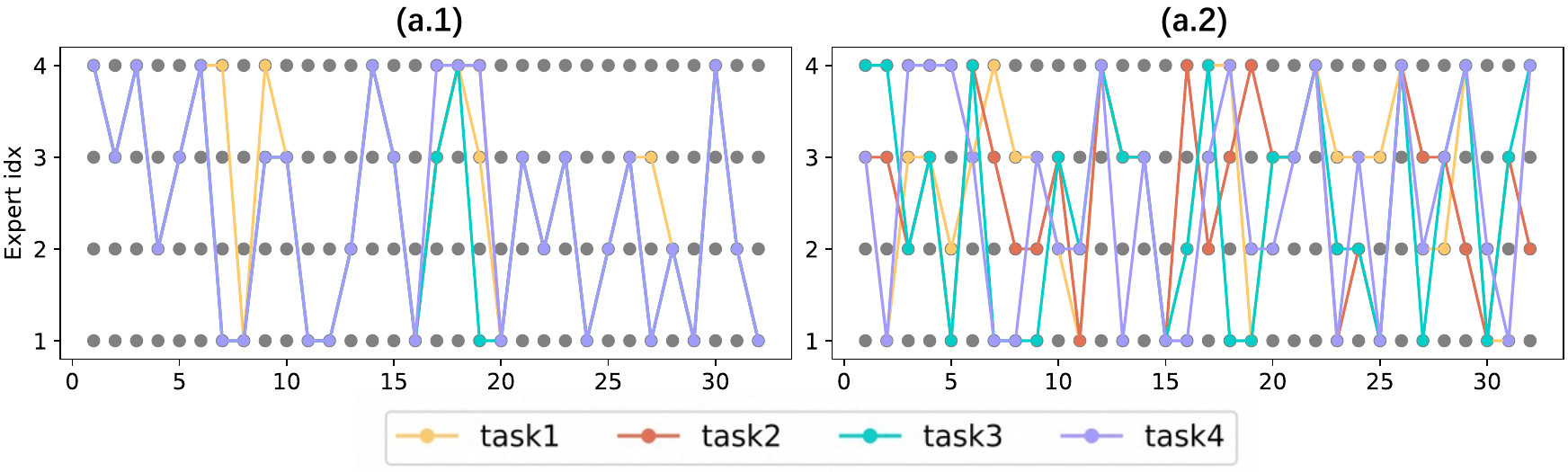}
    \includegraphics[width=0.40\linewidth]{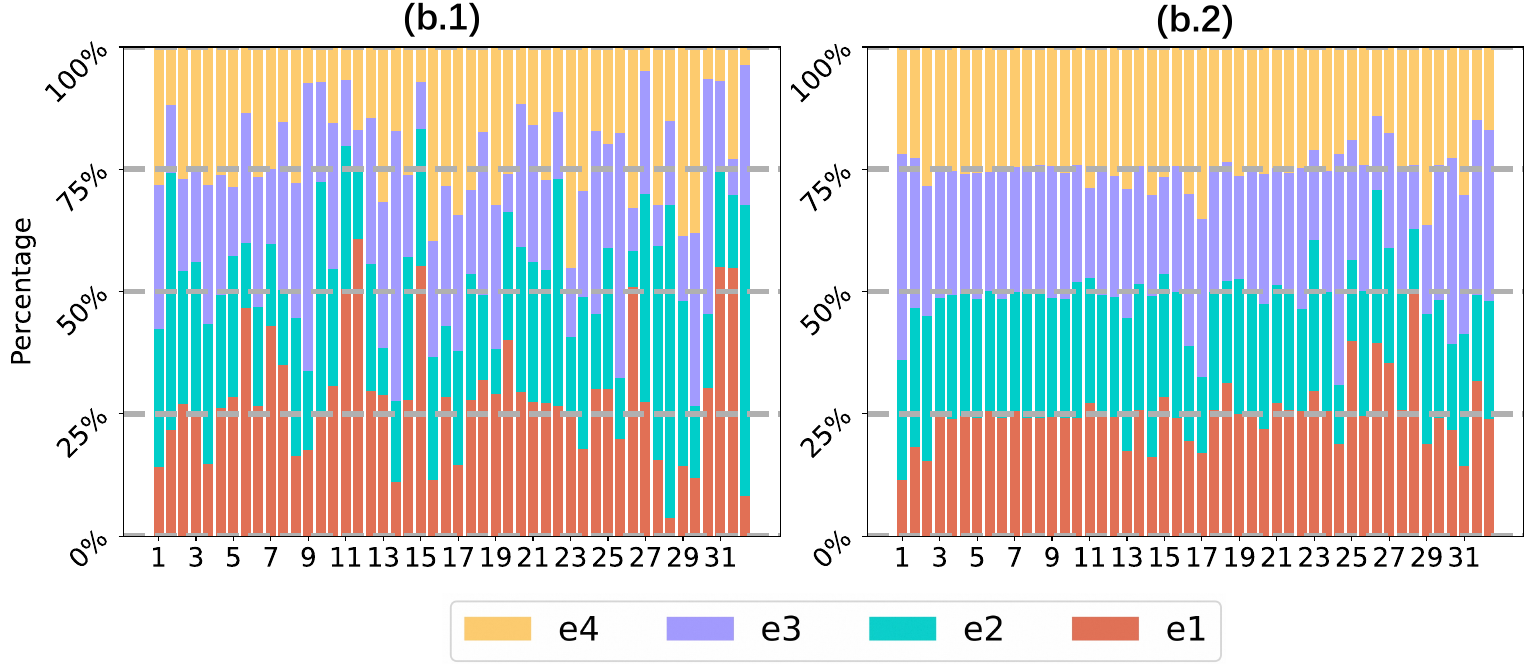}
    \caption{Deep analysis of router. (a) Forward path of expert. (b) Router load visualization.}
    \label{fig:fig5}
\end{figure*}

\begin{table*}
    \centering
    \setlength{\tabcolsep}{1mm}
    \fontsize{9}{11}\selectfont
    % \begin{tabular}{c||c|c|c|c|c|c|c|c|c|c|c||c}
    \resizebox{2\columnwidth}{!}{
    \begin{tabular}{c|c||ccccccccccc||c}
        \Xhline{1.5pt}
        % \multirow{2}{*}{Method} & \makecell{Text\\Summarization} & \multicolumn{2}{c|}{\makecell{Sentiment\\Classification}} & \multicolumn{3}{c|}{NLI} & \makecell{Coreference\\Resolution} &  \multicolumn{4}{c|}{Closed Book QA} & \multirow{2}{*}{Avg.} \\
        % \cline{2-12}
                % \rowcolor[HTML]{DAE0FB}
        %  & OAI-Sum & Emo & IMDB & ANLI & QQP & RTE & WinG & ARC & WQA & NQ & TQA & \\ 
       \multicolumn{2}{c||}{\textbf{Method}} & \textbf{OAI-Sum} & \textbf{IMDB} & \textbf{ANLI} & \textbf{QQP} & \textbf{RTE} & \textbf{WinG} & \textbf{ARC} & \textbf{WQA} & \textbf{NQ} & \textbf{TQA} & \textbf{MMLU} & \textbf{Avg.} \\ 
        \hline
        \hline
        % \rowcolor[HTML]{E0E0E0}
        % Full Fine-Tuning & 28.6 & 90.1 & 96.5 & 60.5 & 88.3 & 88.7 & 74.5 & 72.7 & 52.1 & 26.5 & 39.2& - & 65.25 \\ 
        
         \multirow{3}{*}{Llama-3-8B + }&LoRA$ _{r=32}$& 24.3 & \textbf{95.1} & 47.2 & 78.9 & 80.1 & \textbf{58.3} & \textbf{70.6} & 34.6 & 19.3 & 37.1&\textbf{52.2} & 54.34 \\ 
        &MoELoRA$ _{r=8}$ & \underline{24.8} & \underline{94.9} &\underline{47.6} & \textbf{79.0} & \underline{81.0} & 58.2 & 69.8 & \underline{35.1} & \underline{20.2} & \underline{40.4} &49.2& \underline{54.56} \\ 
        % \hline
        % \cellcolor[HTML]{D9EAD3}
        % LoRAMoG & \textbf{28.0} & \textbf{96.0} & 58.3 & 87.0 & \textbf{88.2} & 72.2 & \textbf{72.2} & 50.6 & 26.1 & 38.5 &42.8& 63.93\\ 
        
        &
        % \cellcolor[HTML]{D9EAD3}
        \textbf{\ourmethod{}}$ _{r=8}$ & \textbf{25.2} & 94.2 & \textbf{49.7} & \textbf{79.0} & \textbf{81.4} & \textbf{58.3} & \underline{70.1} & \textbf{36.1} & \textbf{22.2} & \textbf{41.3}&\underline{52.1}  & \textbf{55.42} \\ 
        \Xhline{1.5pt}
    \end{tabular}}
    \caption{Performance analysis based on different LLM Model.}
    \label{tab:tab3}
\end{table*}

\begin{table*}[h!]
    \centering
    \setlength{\tabcolsep}{1mm}
    \fontsize{9}{11}\selectfont
    % \begin{tabular}{c||c|c|c|c|c|c|c|c|c|c|c||c}
    \resizebox{2\columnwidth}{!}{
    \begin{tabular}{c|c||ccccccccc||c}
        \Xhline{1.5pt}
        % \multirow{2}{*}{Method} & \makecell{Text\\Summarization} & \multicolumn{2}{c|}{\makecell{Sentiment\\Classification}} & \multicolumn{3}{c|}{NLI} & \makecell{Coreference\\Resolution} &  \multicolumn{4}{c|}{Closed Book QA} & \multirow{2}{*}{Avg.} \\
        % \cline{2-12}
        % \rowcolor[HTML]{DAE0FB}
        %  & OAI-Sum & Emo & IMDB & ANLI & QQP & RTE & WinG & ARC & WQA & NQ & TQA & \\ 
       \multicolumn{2}{c||}{\textbf{Method}} & \textbf{MME} & \textbf{MMB} & \textbf{MMB-CN} & \textbf{SEED} & \textbf{POPE} & \textbf{SQA-I} & \textbf{VQA-T} &\textbf{MM-Vet}& \textbf{VizWiz} &\textbf{Avg.} \\ 
        \hline
        \hline
        % \rowcolor[HTML]{E0E0E0}
        % Full Fine-Tuning & 28.6 & 90.1 & 96.5 & 60.5 & 88.3 & 88.7 & 74.5 & 72.7 & 52.1 & 26.5 & 39.2& - & 65.25 \\ 
        
         \multirow{3}{*}{LLaVA-1.5-7B + }&LoRA$ _{r=32}$& \underline{1505.2} & \textbf{62.8} & {53.7} & \textbf{60.2} & \underline{84.8} & 67.8 & 56.9 & \underline{30.2} & 48.4 &\underline{60.01} \\ 
        &MoELoRA$ _{r=8}$ & 1472.7 & 62.3 &\underline{53.8} & 59.5 & 84.4 & \textbf{68.7} & \textbf{57.1} & 30.1 & \underline{48.7} &59.80 \\ 
        % \hline
        % \cellcolor[HTML]{D9EAD3}
        % LoRAMoG & \textbf{28.0} & \textbf{96.0} & 58.3 & 87.0 & \textbf{88.2} & 72.2 & \textbf{72.2} & 50.6 & 26.1 & 38.5 &42.8& 63.93\\ 
        
        &
        % \cellcolor[HTML]{D9EAD3}
        \textbf{\ourmethod{}}$ _{r=8}$ & \textbf{1513.5} & \underline{62.6} & \textbf{54.0} & \underline{60.0} & \textbf{85.3} & \textbf{68.7} &\textbf{57.1} & \textbf{31.2} & \textbf{49.4} &\textbf{60.44}\\ 
        \Xhline{1.5pt}
    \end{tabular}}
    \caption{Performance analysis of MLLM on diverse multimodal benchmarks.}
    \label{tab:tab4}
\end{table*}

\subsection{Overall Performance}
We evaluated the performance of \ourmethod{} in a multi-task learning scenarios using the CME benchmark, compared to other PEFT methods as shown in Table 2. Our observations are summarized as follows: (i) \ourmethod{} (Rank=32) showed the best or second-best performance across multiple tasks, \textbf{with an average score of \textbf{60.29}, significantly higher than other PEFT methods}. Particularly, it achieved the best performance on MMLU, demonstrating \ourmethod{}'s strong capability in handling multi-domain tasks. (ii) Despite a training time of 28 hours for \ourmethod{} (Rank=$16$), slightly longer than baseline methods like LoRA, Prompt-Tuning, and IA3, it achieved competitive average scores of \textbf{59.95} with half the parameter count, highlighting its efficient parameter utilization. (iii) Compared to other multi-LoRA architectures, \ourmethod{} not only showed significant performance improvements but also reduced training costs significantly, with approximately 70\% of MoELoRA and 85\% of HydraLoRA. This demonstrates \ourmethod{}'s effective balance between efficiency and effectiveness.

\subsection{Quantitative Analysis}
\subsubsection{Analysis of Parameter Scales.}
Table 3 explore \ourmethod{}'s performance in multi-task learning across different parameter scales. Experiments demonstrate that \ourmethod{} performs exceptionally well across various parameter configurations, indicating that \ourmethod{} consistently exhibits superior performance compared to MoELoRA. Notably, with an increase in parameter size, LoRA encounters catastrophic forgetting, as evidenced by a sharp decline in scores for TQA (close book QA). In contrast, both MoELoRA and \ourmethod{} alleviate this knowledge collapse, reflecting the stability of their adaptive mechanisms.

\subsubsection{Ablation Analysis.} We conducted an exploration for collaboration and competition modules. As shown in Table 4, both individual modules and their combinations enhance the model's expressive and adaptive capabilities in multi-task scenarios. The collaboration module, utilizing a "\textit{Team}" architecture based on knowledge sharing, effectively promotes the integration and transfer of knowledge among experts, thereby enabling "plug-in" based knowledge organization. The competition module considers the interactions between experts, adjusting the model's preferences for transferring specific knowledge to downstream tasks in response to multi-task performance. The above evidence thoroughly demonstrates the positive significance of the modules.

% \begin{figure*}[h!]
%     \centering
%     \parbox{0.5\textwidth}{
%         \centering
%         \includegraphics[width=\textwidth]{fig/path (1).pdf}
%         \vspace{0.5cm}
%         \includegraphics[width=\textwidth]{fig/path.pdf}
%     }
%     \parbox{0.5\textwidth}{
%         \centering
%         \includegraphics[width=0.48\textwidth]{fig/fig5a.pdf}
%         \hfill
%         \includegraphics[width=0.48\textwidth]{fig/fig5b.pdf}
%     }
%     \caption{整体图示}
%     \label{fig:overall}
% \end{figure*}

% \begin{figure}[h!]
%     \centering
%     \includegraphics[width=0.495\linewidth]{fig/fig2a.pdf}
%     \includegraphics[width=0.48\linewidth]{fig/fig2b.pdf}

%     \caption{\textbf{Time Consumption and Loss Variation.} \textbf{(left)} shows the training time and inference latency relative to LoRA. \textbf{(right)} illustrates the loss variation and its variance range over training time.}
%     \label{fig:fig2}
% \end{figure}

\subsection{In-Depth Analysis}

\subsubsection{Stability Analysis.} In the stability analysis of \ourmethod{}, we examined its performance across different configurations of expert module quantities (see Figure 3(a)). The results indicate that performance improves progressively as the number of expert modules increases from 1 to 4, thanks to the hierarchical knowledge structure and effective      ``plug-in'' knowledge sharing and organization. However, when the number of modules reaches 8, there is a slight decrease in performance, likely due to the added complexity of knowledge transfer with excessive layers. Figure 3(b) illustrates \ourmethod{}'s adaptability to varying data scales, demonstrating its ability to maintain efficient domain knowledge transfer across data scales ranging from 10\% to 100\%, highlighting its potential for multi-task scenarios.

\subsubsection{Computational Costs and Loss Convergence.} Figure 4 illustrates the advantages of \ourmethod{} over MoELoRA in terms of training and inference times. Specifically, TeamLoRA reduces training time by 30\% and increases inference speed by 40\%, as shown in Figure 4(a). Additionally, the loss convergence curve in Figure 4(b) demonstrates that TeamLoRA achieves lower loss values more quickly, highlighting its optimization in training efficiency. 
% These data points clearly indicate a significant improvement in computational costs and training efficiency with TeamLoRA.

\subsubsection{Expert Load Analysis.} We observed the expert paths of MoELoRA across four tasks. The features exhibited overconfidence(see Figure 5(a.1)) in the model's forward path. In contrast, \ourmethod{}, which incorporates a competitive module, effectively learns task-specific models by assigning different expert modules as plug-ins for knowledge combinations(see Figure 5(a.2)). Furthermore, we conducted balanced load testing on 57 tasks in MMLU, as shown in Figure 5(b.1)(MoELoRA) and Figure 5(b.2)(\ourmethod{}). \ourmethod{} demonstrated better load balancing compared to MoELoRA, ensuring greater model stability.

\subsubsection{Performance Comparison of Different Base Models.} To explore the performance of TeamLoRA on other models, we replaced the base model with the more powerful Llama-3 8B and conducted a comprehensive comparison of the CME benchmark. Table \ref{tab:tab3} shows the results of this experiment, where \ourmethod{} consistently demonstrated the best performance. This indicates that \ourmethod{} maintains its advantages in multi-task learning across different base models.

\subsubsection{Performance Analysis of MLLM.} We further expanded the applicability of \ourmethod{} by extending the model from single-modal to multimodal. We fine-tuned the LLaVA-1.5 7B model and evaluated it on nine benchmark tests, including MME~\cite{ref:MME}, MMB/MMB-CN~\cite{ref:MMB}, SEED~\cite{ref:SEED}, POPE~\cite{ref:POPE}, SQA-I~\cite{ref:SQA}, VQA-T~\cite{ref:VQAT}, MM-Vet~\cite{ref:MM-Vet}, and VizWiz~\cite{ref:VizWiz}. As seen, \ourmethod{} achieved the best performance on the majority of benchmarks(see Table \ref{tab:tab4}), indicating that \ourmethod{} demonstrates strong generalizability in multimodal scenarios. Experimental details are provided in Appendix B.

% To assess the impact of varying player counts on cooperation and competition mechanisms, we conducted targeted experiments across all benchmarks. Considering that an increase in player count significantly raises computational costs, we tested configurations with one, two, four, and eight players, with specific performance results presented in Figure \ref{fig:different gamer's number}. The results indicate that configurations with two or four players significantly enhance performance compared to a single-player setup (i.e., LoRA), by moderately introducing cooperation and competition mechanisms. However, when the player count is excessively high, the stability of these mechanisms decreases, leading to performance degradation. Specifically, the two-player and four-player configurations usually exhibited higher scores across all tasks, particularly excelling in Text Summarization and Sentiment Classification tasks, demonstrating sufficient stability and generalization performance. In contrast, the eight-player configuration performed poorly in the Closed Book QA task, reflecting that the introduction of complex cooperative relationships among many players can result in performance that is inferior to that of a single player in certain tasks. These experiments confirm that an appropriate number of players can optimize system performance when designing cooperation and competition mechanisms, while too many players may have the opposite effect, impacting overall effectiveness.

\section{Conclusion}
% In conclusion, this study highlights the efficacy and potential of integrating Parameter-Efficient Fine-Tuning (PEFT) methods with game theory principles through the innovative approach of LoRA with Mixture of Gamers (\ourmethod). By melding Low-Rank Adaptation (LoRA) with Mixture of Experts (MoE) and utilizing game theory-based dynamics, \ourmethod{} significantly advances the field by addressing critical gaps in flexibility and dynamic expert selection inherent in previous methods. The employment of submatrix decomposition alongside Shapley values in \ourmethod{} enables a more granular understanding of the interactions and contributions of different components within PEFT setups. The promising experimental outcomes across a variety of tasks not only underscore \ourmethod's superior performance but also illuminate its versatile applicability and the potential for future adaptations in complex, domain-specific applications. Moving forward, it will be crucial to refine these approaches, ensuring robustness and scalability, to fully harness the transformative power of PEFT in enhancing machine learning models.

\ourmethod{} introduces an innovative PEFT approach by integrating collaborative and competitive modules, which significantly improves the efficiency and effectiveness of multi-task learning. In the proposed CME benchmark tests, \ourmethod{} not only achieves faster response speed but also outperforms existing multi-LoRA architectures in performance. Future research will further explore the game-theoretic framework based on competition and collaboration in multi-LoRA architectures, expanding the potential of PEFT.

\bibliography{ref.bib}

\end{document}